\RequirePackage{fix-cm}
\documentclass[smallextended]{svjour3}       % onecolumn (second format)
\smartqed  % flush right qed marks, e.g. at end of proof
\usepackage[utf8]{inputenc}

\usepackage{amsmath}
\usepackage{amsfonts}
\usepackage{amssymb}
\usepackage{enumitem}
\usepackage{pgf}
\usepackage{tikz}
\usepackage{cite}
\usepackage[nolist, printonlyused]{acronym}

\usepackage{amsthm}

\usepackage[linesnumbered]{algorithm2e}
\usepackage{algpseudocode}
\usepackage{todonotes}
\usepackage{subfig}

\usepackage{xyling}

\usepackage{listings}
\usepackage{xcolor}

\colorlet{punct}{red!60!black}
\definecolor{background}{HTML}{EEEEEE}
\definecolor{delim}{RGB}{20,105,176}
\colorlet{numb}{magenta!60!black}

\usepackage{url}
\usepackage{tabularx}
\usepackage{wrapfig}

\definecolor{lhue}{rgb}{0.1,0.9,0.05}

\definecolor{mygray}{gray}{0.8}

\lstdefinelanguage{scala}{
  morekeywords={abstract,case,catch,class,def,%
    do,else,extends,false,final,finally,%
    for,if,implicit,import,match,mixin,%
    new,null,object,override,package,%
    private,protected,requires,return,sealed,%
    super,this,throw,trait,true,try,%
    type,val,var,while,with,yield},
  otherkeywords={=>,<-,<\%,<:,>:,\#,@},
  sensitive=true,
  morecomment=[l]{//},
  morecomment=[n]{/*}{*/},
  morestring=[b]",
  morestring=[b]',
  morestring=[b]""",
  backgroundcolor=\color{background},
}

\lstdefinelanguage{json}{
    basicstyle=\normalfont\ttfamily,
    numbers=left,
    numberstyle=\scriptsize,
    stepnumber=1,
    numbersep=8pt,
    showstringspaces=false,
    breaklines=true,
    frame=lines,
    backgroundcolor=\color{white},
    literate=
     *{0}{{{\color{numb}0}}}{1}
      {1}{{{\color{numb}1}}}{1}
      {2}{{{\color{numb}2}}}{1}
      {3}{{{\color{numb}3}}}{1}
      {4}{{{\color{numb}4}}}{1}
      {5}{{{\color{numb}5}}}{1}
      {6}{{{\color{numb}6}}}{1}
      {7}{{{\color{numb}7}}}{1}
      {8}{{{\color{numb}8}}}{1}
      {9}{{{\color{numb}9}}}{1}
      {:}{{{\color{punct}{:}}}}{1}
      {,}{{{\color{punct}{,}}}}{1}
      {\{}{{{\color{delim}{\{}}}}{1}
      {\}}{{{\color{delim}{\}}}}}{1}
      {[}{{{\color{delim}{[}}}}{1}
      {]}{{{\color{delim}{]}}}}{1},
}

\newcommand{\baselineSet}{\mathcal{B}}
\newcommand{\solverSet}{\mathcal{S}}
\newcommand{\instanceSet}{\mathcal{I}}

\theoremstyle{definition}

\begin{document}

\begin{acronym}
  \acro{or}[OR]{Operation Research}
\end{acronym}

\title{A Visual Web Tool to Perform What-If Analysis of Optimization Approaches}
\subtitle{}

%\titlerunning{Short form of title}        % if too long for running head

\author{Sascha Van Cauwelaert         \and
        Michele Lombardi \and
        Pierre Schaus
}

%\authorrunning{Short form of author list} % if too long for running head

\institute{S. Van Cauwelaert \at
              Place Sainte Barbe 2 bte L5.02.01 1348 Louvain-la-Neuve \\
              \email{sascha.vancauwelaert@uclouvain.be}           %  \\
           \and
           M. Lombardi \at
              Viale del Risorgimento 2, Bologna (IT)\\
              \email{michele.lombardi2@unibo.it}
          \and
           P. Schaus \at
              Place Sainte Barbe 2 bte L5.02.01 1348 Louvain-la-Neuve \\
              \email{pierre.schaus@uclouvain.be}           %  \\
}

\date{Received: date / Accepted: date}
% The correct dates will be entered by the editor

\maketitle

\begin{abstract}

In Operation Research, practical evaluation is essential to validate the efficacy of optimization approaches. This paper promotes the usage of performance profiles as a standard practice to visualize and analyze experimental results. It introduces a Web tool to construct and export performance profiles as SVG or HTML files. In addition, the application relies on a methodology introduced in \cite{Cauwelaert2015} to estimate the benefit of hypothetical solver improvements. Therefore, the tool allows one to employ what-if analysis to screen possible research directions, and identify those having the best potential. The approach is showcased on two Operation Research technologies: Constraint Programming and Mixed Integer Linear Programming.

% \PACS{PACS code1 \and PACS code2 \and more}
% \subclass{MSC code1 \and MSC code2 \and more}
\end{abstract}

% \newpage
% \setcounter{tocdepth}{3}
% \tableofcontents
% \newpage

\section{Introduction}

In \ac{or}, evaluation is of great importance in order to validate a given solution method with respect to existing ones: for example one may be interested in assessing the effect of an improved neighborhood in Local Search, a faster cut for Mixed Integer Programming, or a global constraint in Constraint Programming.
When reporting research results, it is critical to have the possibility to provide a meaningful analysis and interpretation of tests performed over representative benchmarks.
However, some communities (e.g., Constraint Programming) tend to limit the presentation of the results to tables, sometimes with only a few instances.
This can drastically reduce the significance of the derived conclusions for the general case, which should instead be the primary target when an evaluation is performed.
Finding a meaningful and effective way to aggregate the results is not trivial and it has a direct impact on the conclusions.
Some indicators such as the arithmetic average of normalized measures are well known to bias the results \cite{fleming1986not}.
An additional difficulty is the timeout given to the experiments: some methods indeed become better if they are given more time while others are superior at the early stage of the execution.

A \emph{performance profile} \cite{dolan2002benchmarking} is a tool that is more and more widely employed to grasp a lot of conclusive information out of evaluated benchmarks.
Performance profiles are cumulative distributions for a performance metric that do not suffer from the aforementioned problems.
Among other advantages, they directly provide an approximate cumulative (probability) distribution function\footnote{Under the assumption the studied benchmarks are representative enough.}
that a certain method can solve an arbitrary instance.
Equipped with such a tool, one is clearly able to make better decisions regarding a solver, based on data from quantitative evaluations.

While performance profiles are very useful in practice, there exists no tool to generate and analyze them easily (e.g., via what-if analysis on heterogeneous benchmarks), hampering their broader usage by researchers.
The \ac{or} community would therefore greatly benefit from an easy-to-use Web tool to build and export such profiles, with an easy and well-defined input format.
The current work proposes such a Web-based, free, and public tool.

It is well known that \emph{premature optimization is the root of all evil} \cite{Knuth:1974:SPG:356635.356640}.
It happens that many \ac{or} researchers spend time and energy trying to improve an algorithm that is not the bottleneck of the whole problem.
The tool we introduce permits what-if analysis on the performance profiles.
We can for instance simulate the effect on the whole computation time of reducing the time complexity of a sub-algorithm (e.g., cut generation, global constraint, local search move).

We first describe the tool, and then illustrate how it can be used for the Constraint Programming and the Mixed Integer Linear Programming technologies.

\section{The Public Web Tool}
\label{sec:web_tool}

In order to facilitate the use of performance profiles and hopefully to spread their usage among the community as a standard evaluation tool, we have built a Web tool to construct them easily. This Web tool is publicly available at \url{http://performance-profile.info.ucl.ac.be/}. It allows generating profiles from a simple and well-defined JSON format, and to visualize the approximate effect of improving the solvers considered in an experimentation.

\subsection{Introductory Example}

To describe how to use the tool, we will use a simple running example. While the tool is used to evaluate the performance of solvers, we will use here an analogy with cars. The evaluation results are reported in Table~\ref{table:fictive-results}.
Let us assume that we desire to compare performances of 3 cars, $c_A$, $c_B$ and $c_C$. Those cars will be evaluated according to the time that they need to complete tracks (i.e., from our analogy, the problem instances). The benchmark is made of 6 tracks that have different non-exclusive characteristics that can be used as \emph{labels}. In this example, some tracks can have parts with roads and woods. Finally, the time required by a car to complete a track can be split in several \emph{components}. That is, the total time to complete a track for a car is the sum of the times of all components of the car. For example, the car $c_A$ has a time associated to its wheels and its motor (the rest is considered as negligible). The sum of both time components will be considered as the total time for the car.

\begin{table}[!h]
\centering
\begin{tabular}{ |c|c|c|c|c|c|c| }
  \hline
   Labels & Road & Road  & Road, Wood & Road, Wood & Wood & Wood \\
   \hline
   $c_{A_w}$ & 100 & 30 & 40 & 50 & 10 & 20 \\
     \hline
     $c_{A_m}$ &  20&3&4&5&1&2\\
     \hline
     $c_{B_w}$ & 10&7&45&55&30&50 \\
    \hline
    $c_{C_w}$ & 15&12&35&40&15&25 \\
    \hline
    $c_{C_w}$  & 10&3&4&5&1&2 \\
   \hline
   \end{tabular}
\caption{Results for the introductory car example. $c_{A_w}$ and $c_{A_m}$ stand for the wheels and motor component of $c_A$, respectively.} \label{table:fictive-results}
\end{table}

A \emph{performance profile} is a cumulative distribution of a performance metric for a solver.
Let $\solverSet$ be the set of all considered solvers ($\{c_A, c_B, c_C \}$ in our introductory example) and let $\instanceSet$ be the set of instances (e.g., the set of tracks). We also refer as $\mathit{metric}(s,i)$ to the metric value for the solver $s$ on the instance $i$ (e.g., the time required to complete a track in our example). The profile of the solver $s \in \solverSet$ is then given by:

\begin{align}
F_{s}(\tau) =
\frac{1}{|\instanceSet|}
\left|\left\{ i \in \instanceSet : \frac{\mathit{metric}(s,i)}{\underset{b \in \baselineSet}{\min} (\mathit{metric}(b,i))} \leq \tau \right\}\right|
\label{eq:perf-prof-generic}
\end{align}

where $\baselineSet \subseteq \solverSet$ is the set of baselines, that is, the solvers against which the ratio is computed. If $\baselineSet$ contains all the approaches, the definition is the one of the original work introducing performance profiles \cite{dolan2002benchmarking}. If $\baselineSet$ contains only one solver, it is the definition used in \cite{Cauwelaert2015}. Some intermediary settings are also possible, in an attempt to make the tool as generic as possible.

\begin{figure}[!h]
\centering
\includegraphics[width=\columnwidth]{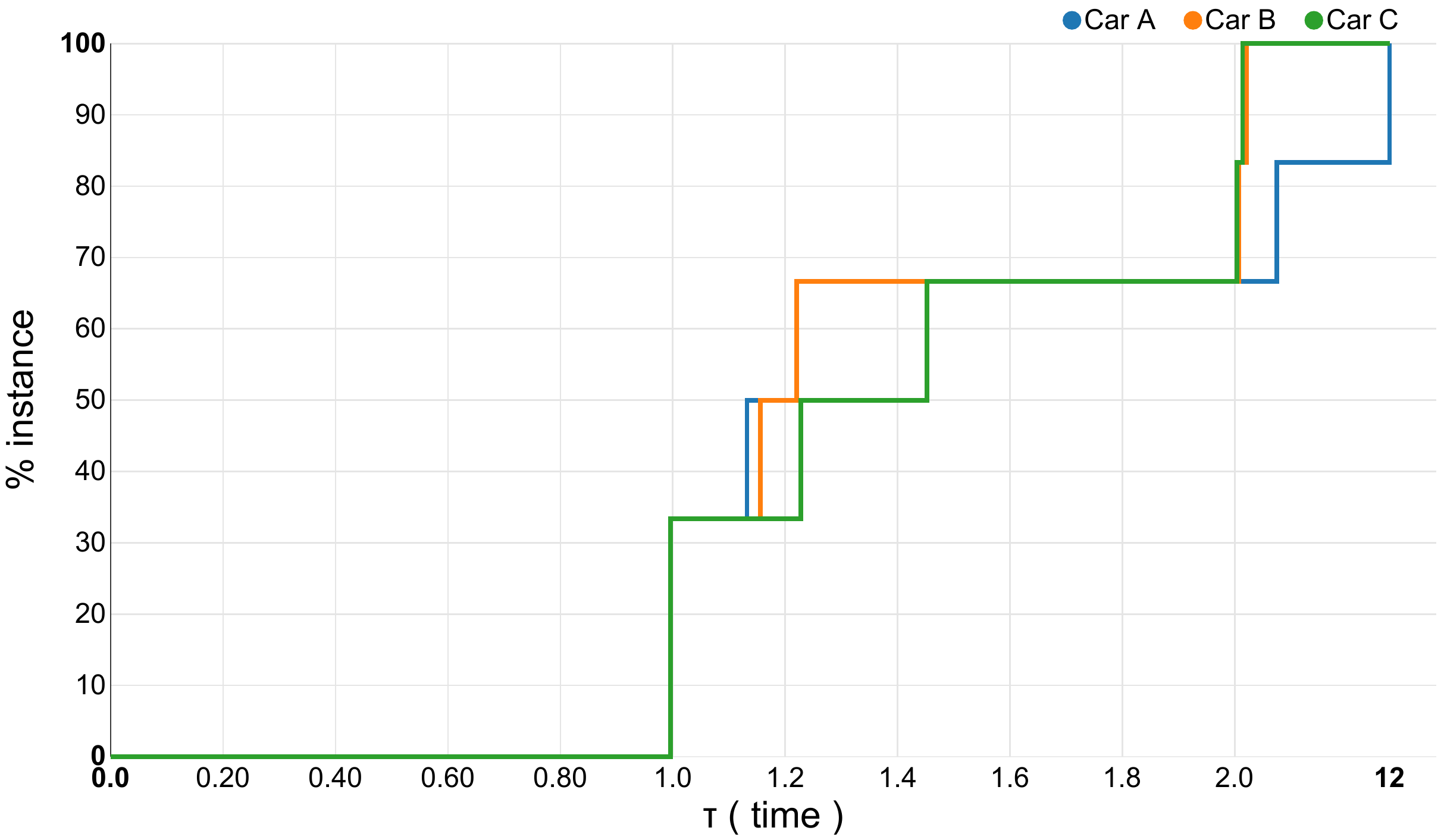}
\caption{Performance profile generated from the data of Table~\ref{table:fictive-results}.}
\label{fig:car-profile}
\end{figure}

When the data from Table~\ref{table:fictive-results} is given to the tool with $\baselineSet = \{c_A, c_B, c_C \}$, the profile given in Figure~\ref{fig:car-profile} is generated. In this figure, one can observe that the x-axis is divided in 2 linear parts. The first one goes from 0 to 2, while the second one goes from 2 to 12. The bounds of the first part can be set by the user (see section \ref{subsec:usage}), in order to select the region of $\tau$ values to be studied. The end of the second part (from 2 to 12 in the figure) cannot be configured and corresponds to the largest metric ratio computed over all the data. This second region provides a shrunk and long-term view of the profiles.

\subsection{Usage}
\label{subsec:usage}

In this section we describe the input and output formats for the Web tool, using our running example. We then describe the different controls that are part of the Graphical User Interface, and their utility.

\begin{figure}[!h]
  \center
  \includegraphics[width=\textwidth]{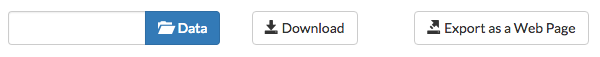}
  \caption{Input as a JSON file and outputs as an SVG file or a Web page.}
  \label{fig:input-output}
\end{figure}

\paragraph{Input/Output:} The input consists of a JSON file that must be passed to the interface (see Figure~\ref{fig:input-output}). It has to be valid according to a JSON schema, available from the Web page. A JSON Schema allows ensuring the JSON file is correct according to the expected format. As an example, Figure~\ref{fig:json-example} shows the JSON file corresponding to the data from Table~\ref{table:fictive-results}:
\begin{itemize}
\item Line 2 defines the \emph{metric} name to be used in the plot legend.
\item Line 3 defines the \emph{labels}, an array of strings that can be associated with each instance separately.
\item Line 4 defines the \emph{instance labels}, an array of arrays. Each element of this array contains the indices of the labels to be associated with the given instance.
\item Lines 5-17 defines the \emph{data} for the different cars, that is, the solvers employed in the experimentation. Each car/solver is defined as a set of components. For a given instance and solver, the total metric is the sum of the metric value for all components.
\end{itemize}

\begin{figure}[!h]

\begin{lstlisting}[language=json,firstnumber=1, escapeinside={(*}{*)}]
{
  "metric" : "time",
  "labels" : ["Road","Wood"],
  "instances" :  [[0],[0],[0,1],[0,1],[1],[1]],
  "data": {
      "Car A": {
        "wheels" : [100,30,40,50,10,20],
        "motor" : [20,3,4,5,1,2]
      },
      "Car B": {
        "wheels" : [10,7,45,55,30,50]
      },
      "Car C" : {
        "wheels" : [15,12,35,40,15,25],
        "motor" : [10,3,4,5,1,2]
      }
  }
}
\end{lstlisting}
\caption{An example of input JSON file for the Web Tool. This file must be correct according to the JSON Schema.} \label{fig:json-example}
\end{figure}

To export the graph, we offer two possibilities (see Figure~\ref{fig:input-output}). First, a button allows downloading the generated profiles as SVG files. SVG is a vector graphics format, similarly to PDF, but that can still be modified if required (using vector graphics tools, or even via a text editor). Moreover, several tools to convert SVG files to PDF exist.
A second button provides the possibility to export the plot as a minimal Web page. This page can then be included by the user in another Web page of his choice.

\paragraph{Controls:}

\begin{wrapfigure}{!h}{0.25\textwidth}
\centering
\includegraphics[width=\linewidth]{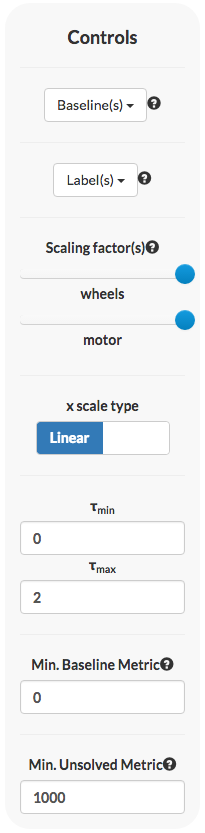}
\caption{\label{fig:controls}The different available controls.}
\end{wrapfigure}

Several controls are available for configuring the plot (see Figure~\ref{fig:controls}). First, the user can decide what is the non-empty set of solvers to be used as baselines.
Moreover, a set of optional labels can be assigned to each instance in order to characterize them (e.g., number of variables, number and type of constraints, authors of the instance, \dots). Some instances might have no label. Using the tool controls, it is possible to specify which labels should be considered in the profile: removing a label from the set will filter out any instance characterized by this label before generating the profiles. This allows investigating easily how the performance profiles are affected by specific groups of instances in the benchmark.

For each \emph{component} in the input data (to which values for the metric are associated), a slider is generated so as to allow the user to study the impact on the performance of being able to reduce the metric for the component by a given ratio. For example, the user may use the slider to quickly assess the impact on the profiles of making wheels that are 20\% more efficient. This feature is very useful for assessing the potential benefits of a certain algorithm improvement, \emph{before actually starting to research how the improvement can actually be obtained}. Alternatively, this technique allows one to estimate the amount of improvement that would be necessary to achieve a given goal.

The minimum and maximum $\tau$ values can be specified via an input box, thus allowing the user to focus on a specific $\tau$ region, or to have a general viewpoint. The button \emph{x scale type} can be used to switch between a linear and a logarithmic scale. If the scale is linear, it is split in two parts: the first one goes from $\tau_\mathit{min}$ to $\tau_\mathit{max}$ and the second one goes from $\tau_\mathit{max}$ to the maximum $\tau$ value computed out of the data. This allows focusing on the desired $\tau$ region (i.e., $\tau_\mathit{min}$ to $\tau_\mathit{max}$) while also having a ``long-term'' view.

Finally, a threshold for the minimum metric value of the baselines can be specified. That is, if one of the approaches considered as a baseline has a value smaller than this threshold for a given instance, this instance is filtered out. This can be used to remove noisy measurements (e.g., imprecisions of time measurements when the solution time is very small) or non-significant measures.
A second threshold for the \emph{minimum unsolved metric} can be specified. That is, if the value of the metric for a solver on a certain instance is larger than this second threshold, the instance will be considered as unsolved. Formally, the metric value for this solver-instance pair is considered infinite. This approach can for example be used to define the time-out value for an experimentation.
% \end{minipage}

\section{Use Cases}
\label{sec:use_cases}
The tool we introduce can be useful for the whole OR community. Let us showcase its usage on two successful technologies, Constraint Programming and Mixed Integer Linear Programming. The different performance profiles we present were constructed using the tool we introduce in this work. We use the definition given in \cite{Cauwelaert2015}, that is, exactly one of the approaches is used as a baseline in Equation~\ref{eq:perf-prof-generic}.

\subsection{Constraint Programming}

Modeling in Constraint Programming is done by specifying a set of constraints to be enforced on a set of variables. The constraints may represent (in principle) any kind of combinatorial relation. To speed-up the search process, each constraint is associated to one or more algorithms that filter out inconsistent values of the variables domains. Those algorithms are generally called \emph{propagators}. Because each propagator $\phi$ is called very often during search, it is important to make its processing time as small as possible. A non-negligible part of research in Constraint Programming aims at improving the time complexity of propagators. However, it is generally an open question whether reducing the time complexity will actually pay off in practice.

We analyzed the Energetic Reasoning propagator for the \textsc{cumulative} constraint\cite{aggoun1993extending,baptiste2000constraint,gay2015simple,gay2015time} on \emph{Resource Constrained Project Scheduling Problems} (RCPSP) instances. This is one of the strongest propagator for \textsc{cumulative} in terms of performed inference, but it comes with a quite high complexity, $\mathcal{O}(n^3)$.

We compare a baseline model to solve the RCPSP with an enriched model that uses the ER algorithm. Moreover, we focus on investigating the potential benefit of having an ER algorithm running in $O(n^2)$ rather than in $O(n^3)$.

The baseline model $M$ employs the Timetabling algorithm from \cite{letort2012scalable} and the ER Checker \cite{baptiste2001constraint}, which both run in $\mathcal{O}(n^2)$ \cite{baptiste2001constraint,derrien2014new}. We did not use the improvements proposed in \cite{derrien2014new}. We use a dynamic search strategy, i.e., the classic \emph{SetTimes} approach from \cite{le1994time}. We consider two benchmarks: the BL instances \cite{baptiste2000constraint} (20-25 activities) and the PSPLIB (j30 and j90, with 30 and 90 activities) \cite{kolisch1999benchmark}. All the implementation was done in the OscaR solver \cite{oscar}.

\begin{figure}[h!]
\centering
\includegraphics[width=\columnwidth]{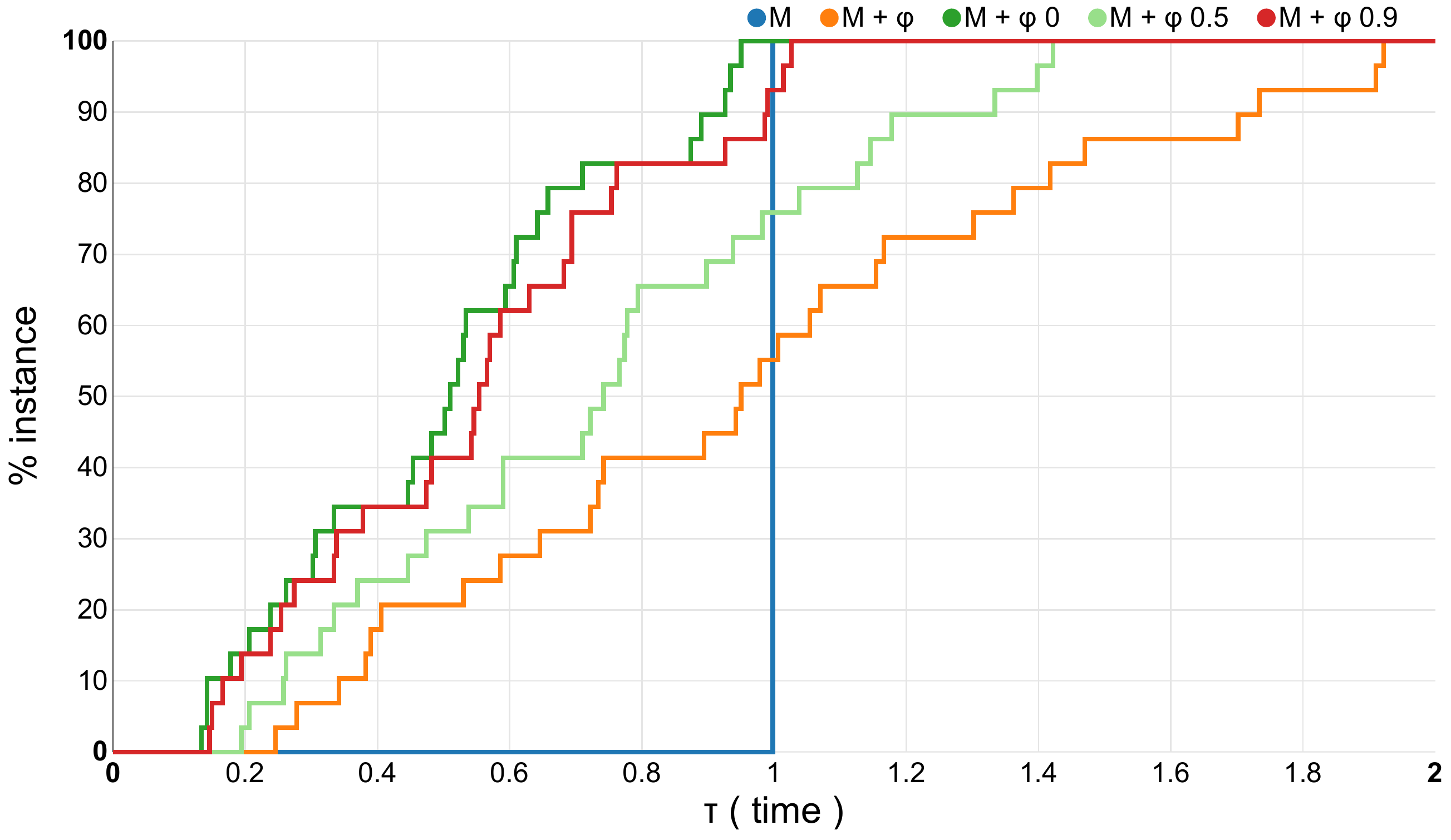}
\caption{Performance profiles for real and fictional ER propagators on the BL instances.} \label{fig:prof-bl}
\end{figure}

\begin{figure}[h!]
\centering
\includegraphics[width=\columnwidth]{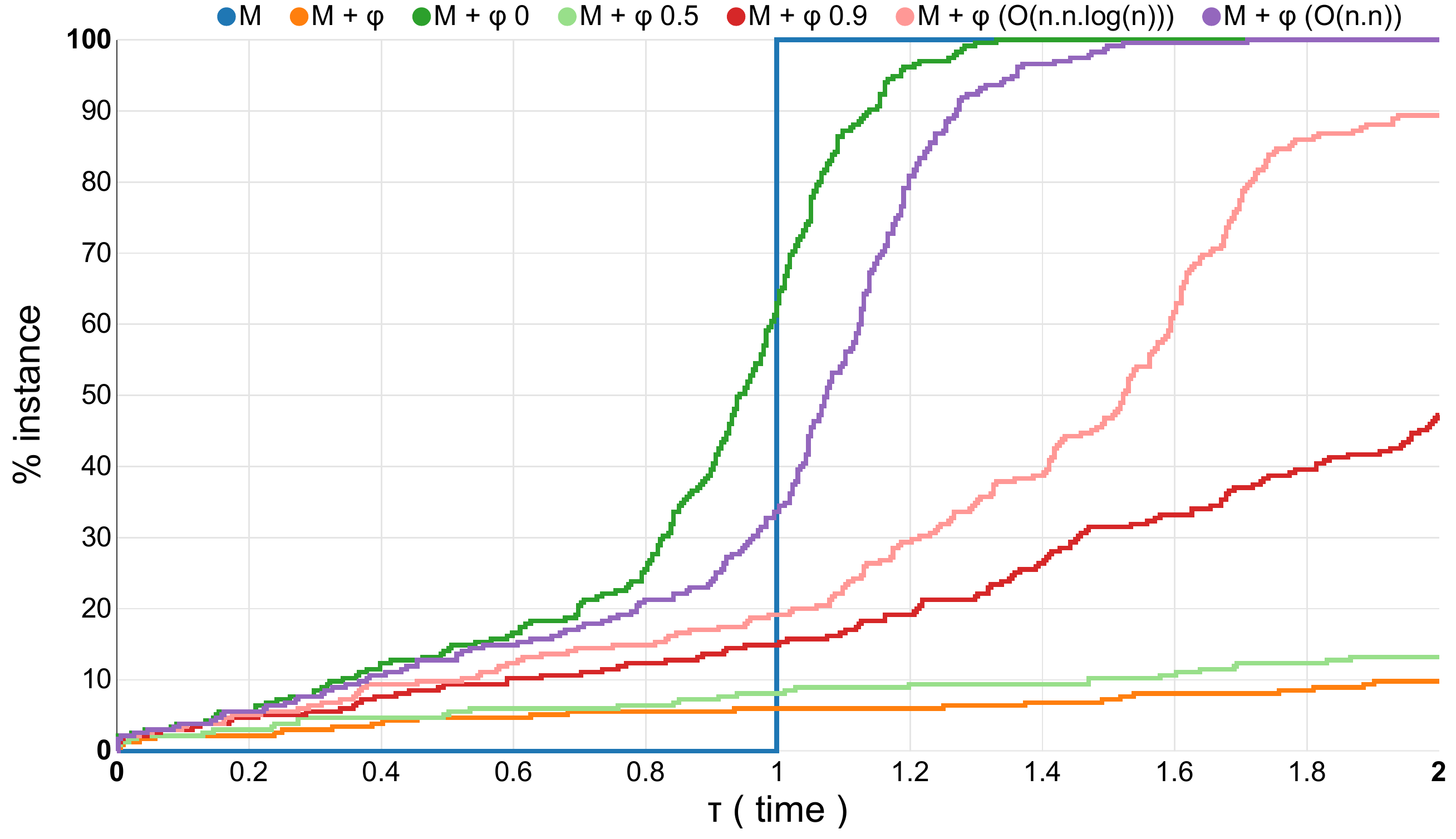}
\caption{Performance profiles for real and fictional ER propagators on the j90 instances.} \label{fig:prof-j90}
\end{figure}

Figure~\ref{fig:prof-bl} and \ref{fig:prof-j90} report profiles respectively for the BL and j90 instances. The real ER propagator provides a more important gain in the case of the BL instances than for the j90 instances. The larger problem size is a likely reason for the performance drop, so it is interesting to analyze the fictional, reduced-cost implementations. In the BL benchmark a cost reduction translates to roughly proportional benefits. On j90, an $O(n^2)$ ER would lead to dramatic performance improvement, but it would beat the baseline in only $\sim35\%$ of the cases. More interestingly, if we consider a null processing time for the ER algorithm (see the dark green curve), one can realize there is about a $35\%$ portion of instances where the baseline would win \emph{no matter what the efficiency of ER is}, i.e., where the additional pruning of ER is sometimes detrimental rather than beneficial. This can be explained by additional calls to the fix point algorithm.

\begin{figure}[h!]
\centering
\includegraphics[width=\columnwidth]{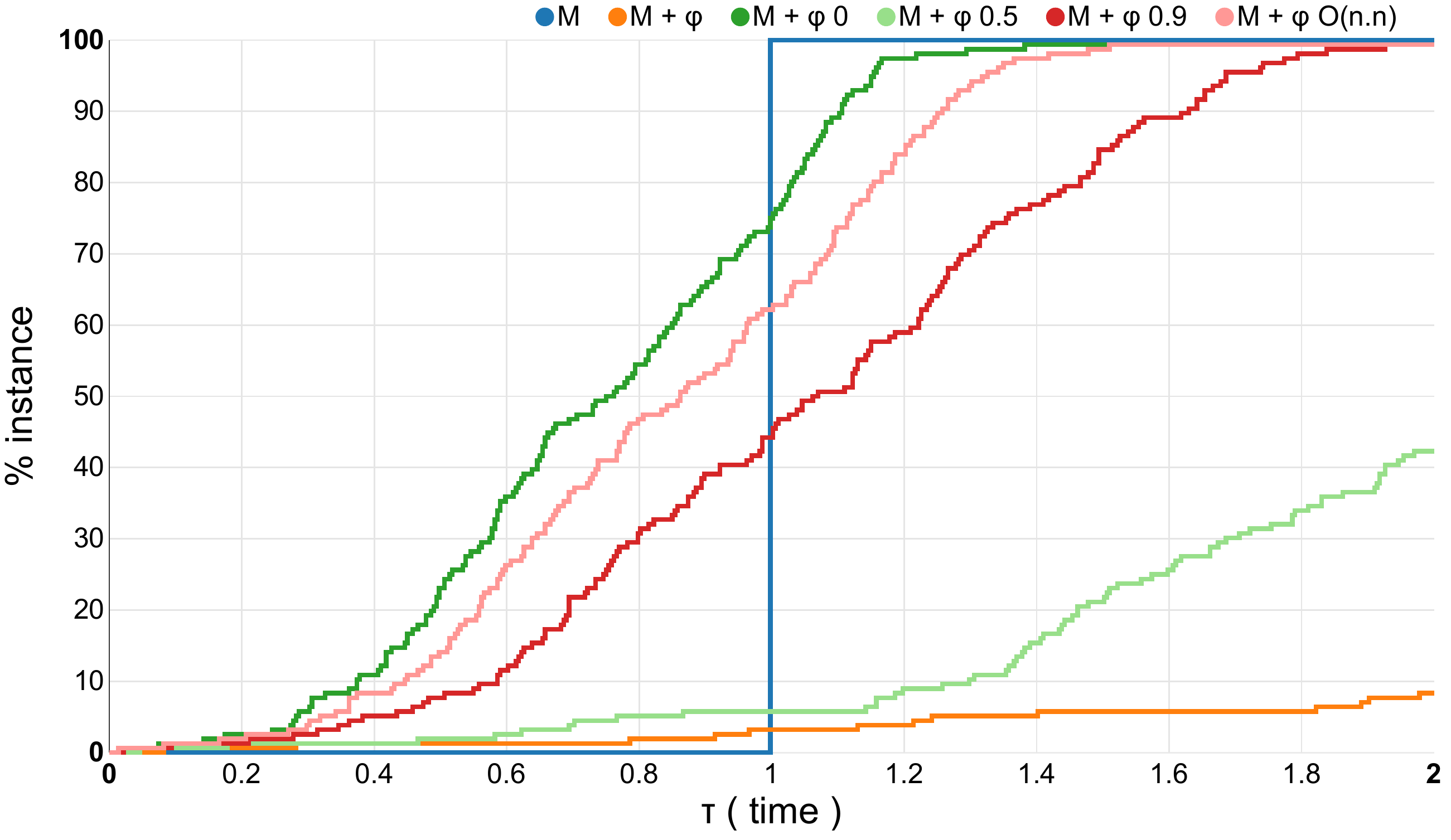}
\caption{Performance profiles for the Binary Static search strategy on the j30 instances.} \label{fig:prof-j30-bs}
\end{figure}

\begin{figure}[h!]
\centering
\includegraphics[width=\columnwidth]{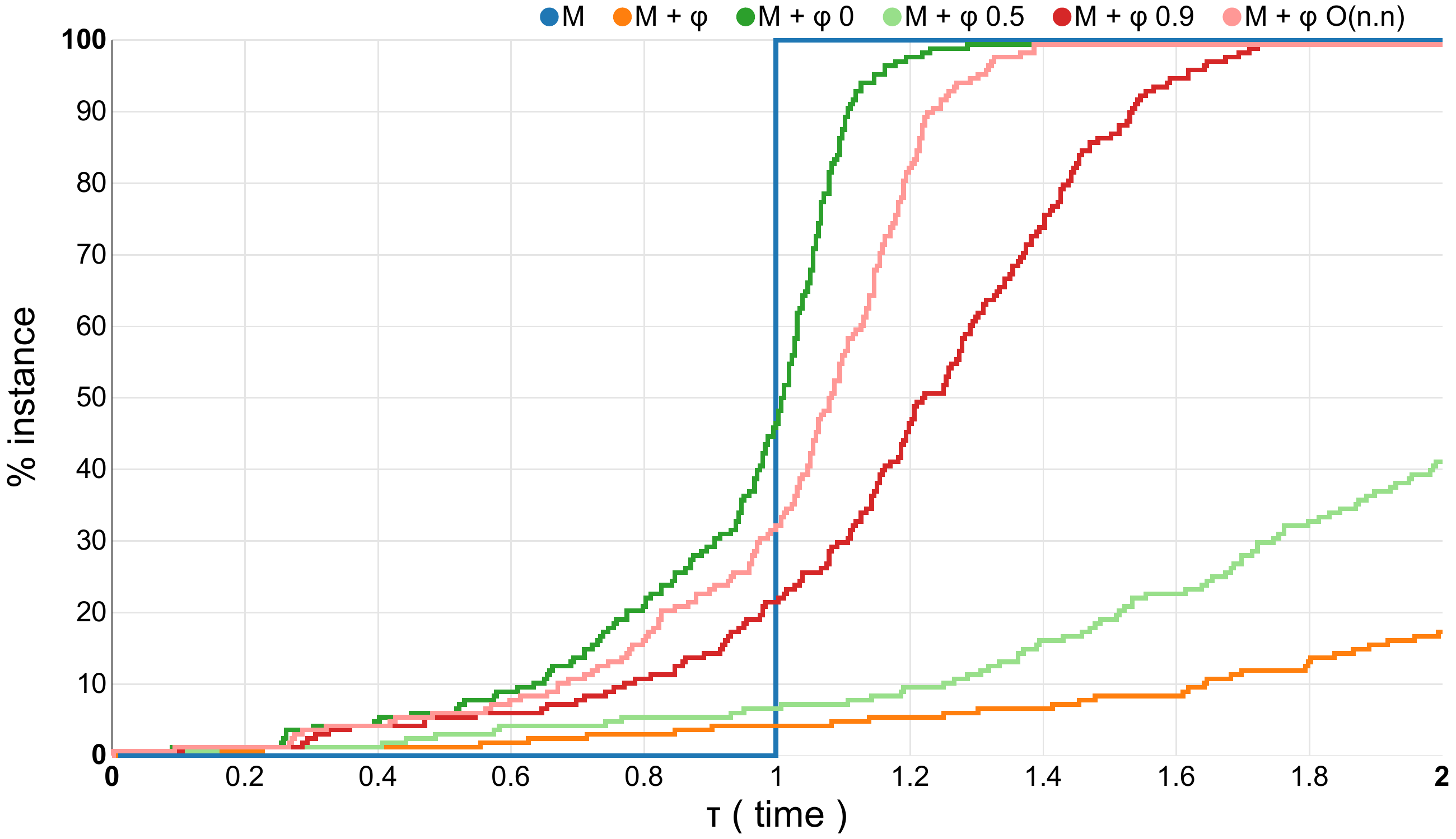}
\caption{Performance profiles the Set Times search strategy on the j30 instances.} \label{fig:prof-j30-st}
\end{figure}

Figures~\ref{fig:prof-j30-bs} and \ref{fig:prof-j30-st} compare profiles for two different search strategies (a lexicographic binary strategy and the \emph{Set Times} strategy \cite{le1994time}) on the j30 instances. The potential gain of reducing the cost is very different for the two strategies, even if the performance of the real propagator is roughly identical. This illustrates the interest of labeling instances in the format.


\subsection{Mixed Integer Linear Programming}

A well-known framework to solve Mixed Integer Linear Programs is the Branch and Cut method, a sophisticated version of Branch-and-Bound. At each node of the search tree, cutting plane algorithms can be used to find additional linear constraints satisfied by all feasible solutions. Those \emph{cuts} sometimes allow discarding important parts of the search space, leading to non-negligible improvements. However, their computation time can be large, so it would be of great interest to know the gain provided by a more efficient computation. Yet if the gain is shown to be negligible, research should focus on other aspects of the solving process.

Let us study the \emph{Concorde} solver \cite{applegate2006concorde}, that has been shown to be efficient at solving the Travelling Salesman Problem. In particular, its authors introduced \emph{local cuts} \cite{applegate2001tsp,applegate2011traveling}, that were crucial to solve some instances. We are interested in knowing the potential of being able to compute those cuts more efficiently.

\begin{figure}[h!]
\centering
\includegraphics[width=\columnwidth]{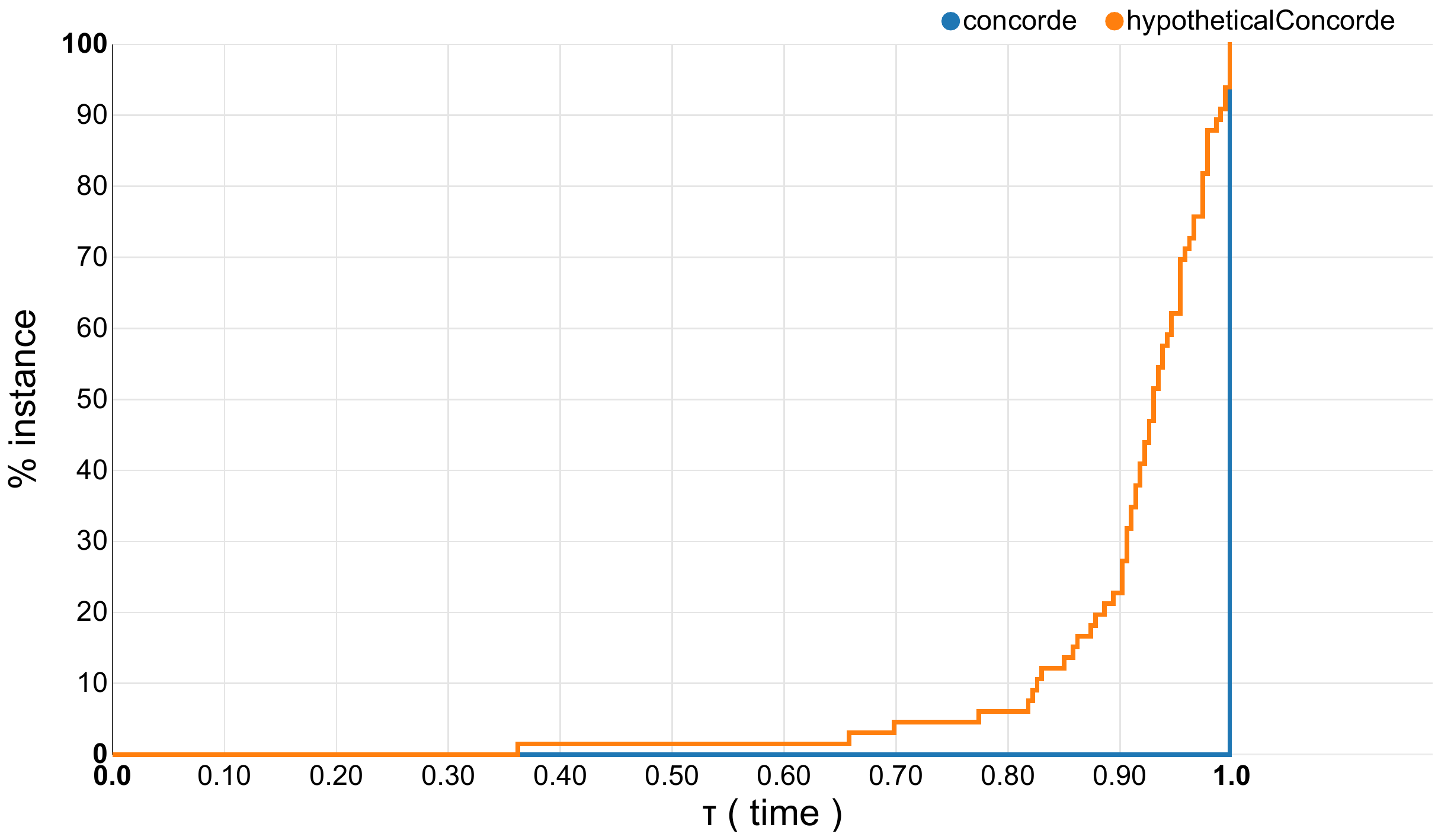}
\caption{Performance profiles for the Concorde Solver on the TSPLib and the VLSI instances.} \label{fig:concorde-tsplib}
\end{figure}

We considered the TSPLib \cite{reinelt1991tsplib} set of instances, augmented with the VLSI\footnote{\url{http://www.math.uwaterloo.ca/tsp/vlsi/index.html}} instances. For each instance, we set a time limit of 900 seconds, and we measured the total time required to compute the local cuts. Instances that could not be solved to optimality or lasted less than 1 sec. were filtered out. Figure~\ref{fig:concorde-tsplib} compares the Concorde solver with an hypothetical version of it, where the local cuts are computed instantly. The profile for this hypothetical solver therefore provides an upper bound of the performance that can be obtained by improving computation of local cuts. For instance, one can see that for $\sim 80\%$ of the instances, at the very most $\sim 15\%$ of the time would be gained by reducing local cuts time computation. This is an indicator that working on this time reduction has a low chance to be fruitful in practice.

% The conclusion we just drove is true for the considered instances, but it might not be not in general, so we generated 200 new instances. Those instances all have 1000 nodes that are randomly distributed uniformly on a square of size 100000. Te distance between the nodes is simply the euclidean distance. Figure~\ref{fig:concorde-svc} gives the obtained result. As one can see, a similar conclusion can be done for those new instances.
%
% \begin{figure}[h!]
% \centering
% \includegraphics[width=\columnwidth]{profiles/concorde/svc.pdf}
% \caption{Performance profiles for the Concorde Solver on the generated instances.} \label{fig:concorde-svc}
% \end{figure}

\section{Conclusion}

In this paper, we introduced a tool that allows researchers of the OR community to easily build and export performance profiles from their experimental data. In addition, we proposed a framework to roughly estimate visually improvements brought to specific parts of a solving process. We believe this methodology can help to avoid falling into pitfalls, where improving the efficacy of a given algorithm used in a particular context is not worth experimentally. We showcased this approach
for propagators in Constraint Programming and cuts in Mixed Integer Linear Programming.

% \clearpage
% \newpage

\end{document}